\documentclass[letterpaper, 10 pt, conference]{ieeeconf}  

\IEEEoverridecommandlockouts                              
\usepackage[pdftex]{graphicx} 

\usepackage{amsthm}
\usepackage{amsmath} 
\usepackage{amssymb}  
\usepackage{bm}  
\usepackage{color}
\usepackage{siunitx}
\usepackage{cite}
\usepackage{balance}

\theoremstyle{plain}
\newtheorem{assumption}{Assumption}
\newtheorem{remark}{Remark}

\begin{document}

\title{\LARGE \bf Variable Inertia Model Predictive Control for Fast Bipedal Maneuvers}
\author{Seung Hyeon Bang$^{1}$, Jaemin Lee$^{2}$, Carlos Gonzalez$^{1}$ and Luis Sentis$^{1}$
 \thanks{$^{1}$S.H. Bang, C. Gonzalez, and L. Sentis are with the Department of Aerospace Engineering and Engineering Mechanics, The University of Texas at Austin, TX 78712, USA
         \tt\small bangsh0718@utexas.edu, lsentis@austin.utexas.edu}%
 \thanks{$^{2}$J. Lee is with the Department of Mechanical and Aerospace Engineering, North Carolina State University, NC 27606, USA
         \tt\small jaemin.lee@ncsu.edu}%
        }

\maketitle

\thispagestyle{empty}
\pagestyle{empty}


\begin{abstract}
This paper proposes a novel control framework for agile and robust bipedal locomotion, addressing model discrepancies between full-body and reduced-order models. Specifically, assumptions such as constant centroidal inertia have introduced significant challenges and limitations in locomotion tasks. To enhance the agility and versatility of full-body humanoid robots, we formalize a Model Predictive Control (MPC) problem that accounts for the variable centroidal inertia of humanoid robots within a convex optimization framework, ensuring computational efficiency for real-time operations. In the proposed formulation, we incorporate a centroidal inertia network designed to predict the variable centroidal inertia over the MPC horizon, taking into account the swing foot trajectories—an aspect often overlooked in ROM-based MPC frameworks. By integrating the MPC-based contact wrench planning with our low-level whole-body controller, we significantly improve the locomotion performance, achieving stable walking at higher velocities that are not attainable with the baseline method. 
The effectiveness of our proposed framework is validated through high-fidelity simulations using our full-body bipedal humanoid robot, DRACO 3~\cite{Bang2023ControlBody}, demonstrating dynamic behaviors.

\end{abstract}
\section{INTRODUCTION}
\label{sec:introduction}

Achieving high-speed, agile locomotion is essential for humanoid robots to execute human-level tasks in dynamic environments. Extensive research has been dedicated to enchancing the stability and robustness of bipedal robots during high-speed motions and stationary phases by optimizing balanced body postures \cite{kim2020dynamic, lee2022online, dai2024multi, Sombolestan2023AdaptiveTerrain,Li2021Force-and-moment-basedRobots}. Given the high-dimensional and nonlinear nature of humanoid robots, many of these studies have simplified the locomotion problem using reduced-order models (RoMs) such as the linear inverted pendulum (LIP) \cite{kim2020dynamic} or single rigid-body (SRB) models \cite{DiCarlo2018DynamicControl}. While these RoMs provide accessible solutions to the locomotion challenges, they often fail to capture critical aspects of robot dynamics, particularly behaviors involving angular momentum. In this paper, we introduce a novel model predictive control (MPC) framework, dubbed Variable Inertia MPC (VI-MPC), which incorporates variable inertia, dependent on the robot's posture, as opposed to the constant inertia typically employed in conventional methods, to enable more stable and agile locomotion.
\vspace{-0.5mm}
\subsection{Related Works}
Model predictive control (MPC) assuming the SRB model has been one of the dominant control strategies applied in quadrupedal locomotion~\cite{ DiCarlo2018DynamicControl, Bledt2017Policy-regularizedCheetah, Kim2019HighlyControl, kim2023safety}. More recently, this approach has been adapted in humanoid robots~\cite{Garcia2020CentroidalMPC, Li2021Force-and-moment-basedRobots, Romualdi2022OnlineAdjustment, Ding2022Orientation-AwareWalking, Rossini2023AAvoidance}. A commonality across these applications is their focus in optimizing locomotion variables such as reaction forces or footsteps under the assumption that the robot can be modeled as a single lumped mass—often informally referred to as the ``potato model." However, in stark contrast to most quadruped robots, this simplification falls short for humanoid robots that feature heavy legs. Specifically, significant changes in the rotational inertia tensor, e.g., resulting from leg swings, compromise the model's accuracy. Since the MPC's performance is highly dependent on the precision of the underlying model, the SRB model-based MPC controller becomes less suitable for humanoid robots undergoing high inertia variations. 

\begin{figure}[t!]
    \centering
    \includegraphics[width=\columnwidth]{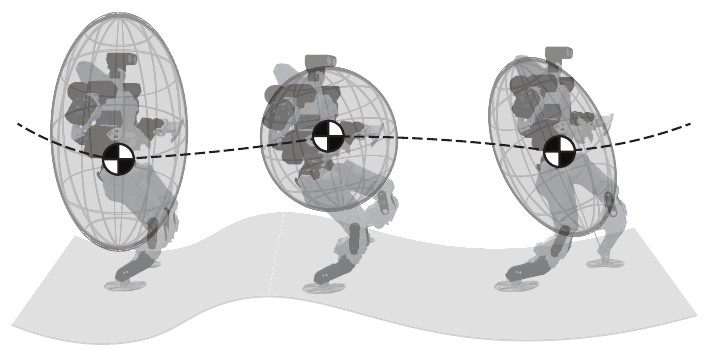}
    \caption{\textbf{Illustration of the Variable Inertia MPC (VI-MPC framework)}: VI-MPC plans using the SRB in conjunction with variable centroidal inertia to minimize model discrepancies when compared to the full-body model.}
    \label{fig:velocity_tracking}
    \vspace{-5mm}
\end{figure}
It is nontrivial to reason about the centroidal inertia and to incorporate it in the ROMs without considering the joint configurations. Recent works~\cite{Wang2021ARobots, ahn2021towr+} proposed methods to augment the SRB model with time-varying centroidal inertia by either abstracting a leg as a point mass or creating a neural network (NN) trained with a full-kinematics model. While the augmented SRB models are capable of generating dynamic and versatile motions, these methods rely on offline trajectory optimization, which poses computational challenges. In response, Garcia et al.~\cite{Garcia2020CentroidalMPC} presented a linear MPC constructed with a system that linearizes the rotational dynamics, taking into account the angular momentum effects induced by the movement of each link. However, it is not capable of explicitly integrating the changing inertia effects during leg swings, since it does not consider the swing foot trajectory. Morevoer, it relies on predefined angular momentum reference trajectories, which are challenging to design intuitively.

Alternatively, recent works have explored the use of online nonlinear MPC for generating whole-body motions that account for significant changes in angular momentum by employing more expressive models. For example,~\cite{Meduri2022BiConMP:Planning} utilizes the centroidal dynamics and whole-body kinematics optimizers to compute the centroidal and joint trajectories. Similarly,~\cite{Mastalli2022AgileApproach} employs a full-body dynamics model to explicitly consider the nonholonomic nature of angular momentum. While these approaches have demonstrated effectiveness, they still suffer from slow computation times and susceptibility to local minima.

\subsection{Contributions}
This paper proposes the VI-MPC framework that accounts for variations in centroidal rotational inertia resulting from leg swing motions. Building on our previous work in \cite{ahn2021towr+}, we utilize a centroidal composite inertia neural network (CCINN) offline to obtain variable inertial properties. These properties are then seamlessly integrated into our convex MPC formulation to plan locomotion trajectories, leveraging both whole-body orientation (WBO) and the SRB model. The generated trajectories are subsequently tracked using an optimization-based whole-body controller (WBC).

The main contributions of this paper are threefold. First, by incorporating variable inertial properties and WBO into the proposed framework, our approach captures more precise angular dynamics, leading to improved accuracy over conventional methods that rely solely on base link orientation~\cite{Garcia2020CentroidalMPC,Li2021Force-and-moment-basedRobots, Ding2022Orientation-AwareWalking}. Second, the combination of offline and online processes in the proposed framework ensures more stable and efficient computations, minimizing the risk of converging to local minima during the optimization process. Lastly, we demonstrate that the proposed MPC approach facilitates more dynamic locomotion, including longer strides, higher walking speeds, and more agile turning behaviors, by effectively bridging the gap between the full-body and SRB models.

The organization of the remainder of this paper is as follows: In Section~\ref{sec:preliminary}, we revisit the dynamics models for control and formally define our problem. Section~\ref{sec:proposed_approach} details our approach to solving this problem through the MPC and WBC framework. Then, we present a set of numerical simulation results on the MPC in Section~\ref{sec:experiment_results}. Lastly, Section~\ref{sec:conclusion} concludes the paper with potential directions for future work.



\section{PRELIMINARIES}
\label{sec:problem_statement}
\label{sec:preliminary}
\begin{figure*}[t!]
    \centering
    \includegraphics[width=\textwidth]{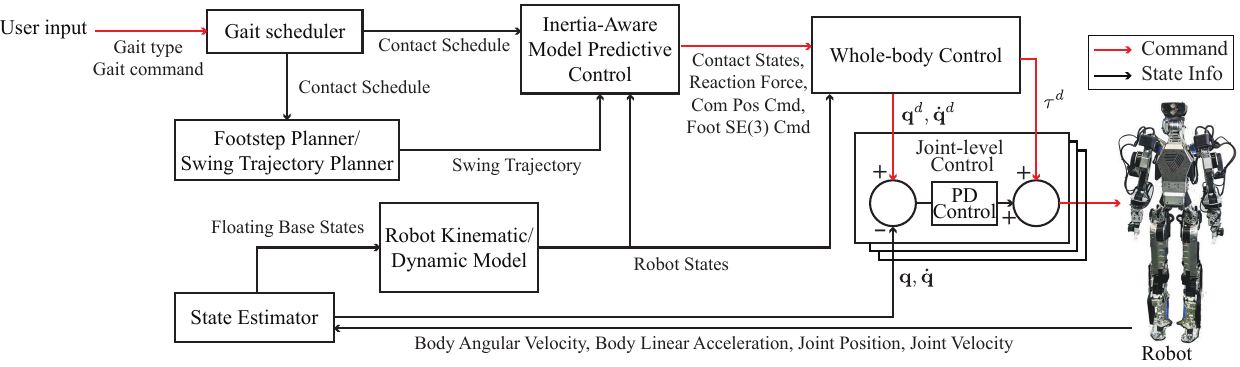}
    \caption{\textbf{The Proposed Hierarchical Control Framework.} The user input commands the gait type, speed, and direction. The model predictive controller then calculates the desired reaction forces, CoM position, whole-body orientation, and foot SE(3) commands. Finally, the whole-body controller computes joint position, velocity, and torque commands, which are sent to each joint-level controller.}
    \label{fig:mpc_framework}
    \vspace{-5mm}
\end{figure*}

In this section, we lay the foundations by introducing the dynamic models employed in our hierarchical control framework designed for dynamic locomotion of humanoids. Subsequently, we outline our problem statement and our proposed solution.    
\subsection{Full-body Dynamic (FBD) Model}
The rigid body dynamics equation of a floating base robot with $n$ degrees of freedom (DoF), actuated by $n_a$ joints ($n_a < n$), is given by:
\begin{align} \label{eq:multi-body dynamics}
    \mathbf{M}(\mathbf{q}) \dot{\mathbf{v}} + \mathbf{b}(\mathbf{q}, \mathbf{v}) = \mathbf{S}_a^{\top} \bm{\tau} + \mathbf{J}_c(\mathbf{q})^{\top} \mathbf{F}_r,
\end{align}
where $\mathbf{q} = [\mathbf{q}_f^{\top}, \mathbf{q}_a^{\top}]^{\top} \in \mathcal{Q}$, denotes the robot's configuration, $\mathbf{q}_f \in SE(3)$ represents the position and orientation of the system's floating base, $\mathbf{q}_a \in \mathbb{R}^{n_a}$ is the actuated joints' configuration, and $\mathbf{v} \in \mathbb{R}^{n_a + 6}$ the system's generalized velocity. $\mathbf{M}(\mathbf{q}) \in\mathbb{R}_{>0}^{(n_a + 6)\times (n_a + 6)}$, and $\mathbf{b}(\mathbf{q},\mathbf{v})\in\mathbb{R}^{n_a + 6}$ denote the mass/inertia matrix and the sum of the centrifugal/Coriolis and gravitational force, respectively.  The selection matrix for the actuated joint torque vector $\bm{\tau}\in\mathbb{R}^{n_a}$ is represented by $\mathbf{S}_a\in\mathbb{R}^{n_a\times (n_a + 6)}$. In addition, $\mathbf{F}_r \in \mathbb{R}^{n_c}$ represents the concatenation of the $6$-dimensional contact wrenches, and $\mathbf{J}_c(\mathbf{q}) \in \mathbb{R}^{n_c \times (n_a + 6)}$ is the corresponding contact Jacobian. For simple notations, we use $\mathbf{M}$, $\mathbf{b}$, and $\mathbf{J_c}$ instead of using $\mathbf{M}(\mathbf{q})$, $\mathbf{b}(\mathbf{q}, \dot{\mathbf{q}})$, and $\mathbf{J_c(\mathbf{q}})$, respectively.  

The dynamics equation in~\eqref{eq:multi-body dynamics} can be decomposed into its unactuated and actuated parts as follows~\cite{Herzog2016StructuredGeneration}:
\begin{subequations}
    \begin{align}
        \mathbf{M}_u\dot{\mathbf{v}} + \mathbf{b}_u &=  \mathbf{J}_{c,u}^{\top} \mathbf{F}_r, \\
       \mathbf{M}_a\dot{\mathbf{v}} + \mathbf{b}_a &=  \bm{\tau} + \mathbf{J}_{c,a}^{\top} \mathbf{F}_r 
    \end{align}
\end{subequations}
where the subscript $u$ and $a$ correspond to unactuated and actuated dynamics, respectively. The unactuated dynamics is equivalent to the robot centroidal momentum dynamics as follows:
\begin{align} \label{eq:centroidal_momentum}
   \begin{bmatrix}
   \mathbf{\dot{l}}_G \\
   \mathbf{\dot{k}}_G
   \end{bmatrix} =
   \begin{bmatrix}
       m\mathbf{g} + \sum_{i=1}^{n} \mathbf{f}_i \\
       \sum_{i=1}^{n} \mathbf{r}_i  \times \mathbf{f}_i + \bm{\tau}_i 
   \end{bmatrix}
\end{align}
where $\mathbf{l}_G$ and $\mathbf{k}_G$ denote the linear and angular momentum expressed at the CoM. $m$ represents the lumped mass of the robot, and $\mathbf{g}\in \mathbb{R}^{3}$ is the gravitational acceleration. $\mathbf{f}_i\in \mathbb{R}^{3}$ 
and $\bm{\tau}_{i}\in \mathbb{R}^{3}$ represent the $i^{\textrm{th}}$ ground reaction force and torque, respectively. $\mathbf{r}_i\in \mathbb{R}^{3}$ is the $i^{\textrm{th}}$ footstep position relative to CoM. Note that all the quantities are expressed in the inertial frame unless explicitly stated otherwise.

\subsection{Single Rigid Body Dynamic Model} \label{sec:srb}
ROMs offer the advantage of capturing the simplicity of locomotion principles while enabling the stabilization of robotic systems in fast and reactive manners. The SRB model has been among the most widely used to perform legged robot locomotion. The SRB model includes three major assumptions: 1) the leg dynamics have a negligible impact on the centroidal states of the robot, 2) the utilization of a constant rotational inertia matrix, $\mathbf{I}\in\mathbb{R}^{3\times3}$, calculated based on the robot's nominal configuration, and 3) the dynamics of the robot's base orientation fully capture its overall orientation dynamics. Applying these assumptions to the centroidal momentum dynamics in~\eqref{eq:centroidal_momentum}, the equations of motion of the SRB model and its rotational motion are expressed as:
\begin{align}
    m\ddot{\mathbf{p}}_c &= m\mathbf{g} + \sum_{i=1}^{n} \mathbf{f}_i, \label{eq:linear_momentum}\\
   \frac{d}{dt}(\mathbf{I}\bm{\omega})  &= \sum_{i=1}^{n} \mathbf{r}_i  \times \mathbf{f}_i + \bm{\tau}_i,   \label{eq:angular_momentum} \\
   \dot{\mathbf{R}} &= \bm{\omega} \times \mathbf{R}, \label{eq:rotation_kinematics}
\end{align}
where $\mathbf{p}_c\in \mathbb{R}^{3}$ denotes the position of the robot's Center of Mass (CoM). $\mathbf{R} \in SO(3)$ and $\bm{\omega} \in \mathbb{R}^{3}$ denote the rotation matrix, which transforms from the body frame to world coordinates, and the body angular velocity.

\subsection{Whole-body Orientation} \label{sec:wbo}
To represent the orientation of multibody-legged robots, whole-body orientation (WBO)~\cite{ChenIntegrableRobots}, which depends on the robot's configuration rather than a history of motion, is given by:
\begin{align}
   \mathbf{Q}_{B, WBO,x,y,z}(\mathbf{q};\bm{\Theta}) = \bm{\Theta}\bm{\lambda}(\mathbf{q}), \\
   \mathbf{Q}_{WBO} =\mathbf{Q}_{B} \otimes \mathbf{Q}_{B, WBO}
\end{align}
where $\mathbf{Q}_{B, WBO} \in \mathbb{H}$ denotes the WBO frame orientation relative to the base, represented using quaternion notation. $\mathbf{Q}_{B, WBO, x,y,z}$ represents the $x$, $y$, $z$ components of the quaternion. $\otimes$ denotes quaternion multiplication. Here, $\bm{\lambda}(\mathbf{q})$ and $\bm{\Theta} \in \mathbb{R}^{3 \times n_\lambda}$ are a vector of basis function with dimension of $n_{\lambda}$ and its coefficient matrix, respectively. $\mathbf{Q}_{WBO} \in \mathbb{H}$ and $\mathbf{Q}_{B} \in \mathbb{H}$ represent the orientation of the WBO frame in the inertial frame, and the robot's base frame in the inertial frame, respectively. The whole-body angular velocity, equivalent to the rotational part of the inertia-weighted average spatial velocity~\cite{Orin2013CentroidalRobot}, is given by:
\begin{align}
   \bm{\omega}_{WBO} = \mathbf{I}_G^{-1}\mathbf{k}_G 
\end{align}
where $\mathbf{I}_G$ represent the the centroidal composite rigid body inertia (CCRBI)~\cite{Lee2007ReactionRobots}.

\subsection{Problem Statement}
As mentioned previously, ROMs come with several assumptions that may not be suitable for full-body humanoid robots. Especially, the behavior of each limb can exert a substantial influence on the centroidal dynamics, rendering ROMs inadequate in capturing such complicated dynamics. Furthermore, the limited consideration of varying rotational inertia with respect to the robot's configuration poses significant challenges. More specifically, this limitation results in notable model discrepancies between the full-body model and ROMs, especially when dealing with angular motions. 

This paper addresses a locomotion challenge in full-body humanoid robots arising from state-dependent rotational inertia while executing locomotion. In contrast to conventional ROM-based approaches, we consider the impact of angular motions in our framework. More specifically, we formulate the convex MPC problem to reduce the discrepancy between the full-body and SRB models by considering the variable rotational inertia, mostly induced by foot swinging. The MPC, in turn, generates more dynamically consistent contact wrenches. Furthermore, our whole-body control scheme allows precise control of multiple tasks based on the results derived from the proposed MPC. The proposed methods enhance the robustness of our framework executing dynamic locomotion.

\section{THE PROPOSED APPROACH}
\label{sec:solution_approach}
\label{sec:proposed_approach}
This section introduces our proposed variable inertia MPC-based control framework designed to enable robust dynamic locomotion of humanoid robots, as depicted in Fig~\ref{fig:mpc_framework}. We begin this section by introducing the simplified model used in our framework (\ref{sec:proposed_approach}-A). Nest, we describe the CCIN (\ref{sec:proposed_approach}-B) and the process for generating the reference trajectory (\ref{sec:proposed_approach}-C). Following that, we explain the proposed convex MPC formulation (\ref{sec:proposed_approach}-D) and the WBC desgined to trck the results of the MPC approach (\ref{sec:proposed_approach}-E). 

\subsection{Simplified Model}
Inspired by the work of quadruped locomotion control~\cite{DiCarlo2018DynamicControl, Villarreal2020MPC-basedLocomotion}, we model the robot as an SRB subject to contact patches at each stance foot. However, there are two key differences in our approach: first, we explicitly take the full-body composite inertia into account along with the SRB. Additionally, instead of the robot's torso orientation, we employ the WBO dynamics for the SRB orientation dynamics with the following assumption: 
\begin{assumption}
    In our problem, we assume negligible precession and nutation of the rotating body as in~\cite{DiCarlo2018DynamicControl}, namely $\bm{\omega}\times\mathbf{I}\bm{\omega} \approx 0$.  
\end{assumption}
\noindent
With the above assumption, we rewrite the SRB rotational dynamics equations \eqref{eq:angular_momentum} and \eqref{eq:rotation_kinematics} in \ref{sec:srb} as follows:
\begin{align}
   \mathbf{I}(\mathbf{q})\dot{\bm{\omega}}_{\textrm{WBO}}  &= \sum_{i=1}^{n} \mathbf{r}_i  \times \mathbf{f}_i + \bm{\tau}_i, \label{eq:angular_momentum_wbo}\\
   \dot{\mathbf{R}}_{\textrm{WBO}} &= \bm{\omega}_{\textrm{WBO}} \times \mathbf{R}_{\textrm{WBO}} \label{eq:wbo_kinematics}
\end{align}
where $\mathbf{I}(\mathbf{q})\in\mathbb{R}^{3\times3}$ denotes the centroidal composite rigid body inertia (CCRBI)~\cite{Orin2013CentroidalRobot}, which is a function of joint configuration $\mathbf{q}$. $\mathbf{R}_{\textrm{WBO}}$ represents the rotational centroidal orientation of a multibody system, while $\bm{\omega}_{\textrm{WBO}} \in \mathbb{R}^3$ denotes the system's whole body angular velocity. Given $\mathbf{T}(\bm{\Theta}) \in \mathbb{R}^{3\times3}$, which is the matrix that maps from Euler angle rates to angular velocities, we introduce an assumption as follows:
\begin{assumption}
    In our problem, we assume the matrix $\mathbf{T}$ is always invertible, namely $\phi \neq \frac{\pi}{2}$ (robot's WBO is not pointed vertically, which in practice does not happen).
\end{assumption}
\noindent
Then, for state-space representation, the rotational kinematics in~\eqref{eq:wbo_kinematics} is rewritten using Euler angles as follows:
\begin{align}
    \dot{\bm{\Theta}}_{\textrm{WBO}} = \mathbf{T}^{-1}(\bm{\Theta}_{\textrm{WBO}})\bm{\omega}_{\textrm{WBO}}, \label{eq:euler_angle}
\end{align}
where $\bm{\Theta}_{\textrm{WBO}}= [\theta, \phi, \psi]$ follows an earth-fixed roll($\theta$), pitch($\phi$), yaw($\psi$) convention.

\begin{remark}
At each MPC control frequency, the orientation states of the SRB model are updated based on the robot's whole-body orientation (WBO) and its whole-body angular velocity (inertia-weighted average spatial velocity) rather than from the torso's orientation and angular velocity, as is traditionally done.
\end{remark}
We define the state variable and control input as follows:
\begin{equation}
    \begin{split}
        \bm{x} =&[\bm{\Theta}_{\textrm{WBO}}^{\top}, \bm{p}_c^{\top}, \bm{w}_{\textrm{WBO}}^{\top}, \dot{\bm{p}}_c^{\top}]^{\top}  \in \mathbb{R}^{n_x} ,\\
        \bm{u} =& [\bm{\tau}_1^{\top}, \bm{f}_1^{\top}, \cdots, \bm{\tau}_n^{\top}, \bm{f}_n^{\top} ]^{\top}\in\mathbb{R}^{n_u}.
    \end{split}
\end{equation}
After rearranging~\eqref{eq:linear_momentum},~\eqref{eq:angular_momentum_wbo}, and~\eqref{eq:euler_angle} usng the defined state and input variables,  
the state space model is written as follows:
\begin{equation}
\begin{aligned}
    \dot{\bm{x}}(t) &= \mathbf{A}(\bm{\Theta}_{\textrm{WBO}})\bm{x}(t) + \mathbf{B}(\mathbf{I}(\mathbf{q}), \bm{\Theta}_{\textrm{WBO}}, \bm{r}_{1:n})\bm{u}(t) + \mathbf{g} \\
    &= \left[ 
    \begin{array}{c}
        \mathbf{T}^{-1}(\bm{\Theta}_{\textrm{WBO}})\bm{w}_{\textrm{WBO}} \\
        \dot{\bm{p}}_c \\
        \mathbf{I}^{-1}(\mathbf{q}) \sum_{i=1}^{n}\mathbf{\Phi}_i\left([\bm{r}_i]_\times\mathbf{f}_i + \mathbf{\bm{\tau}}_i\right) \\
        \frac{1}{m}\sum_{i=1}^{n} \mathbf{\Phi}_i\mathbf{f}_i + \mathbf{g},
    \end{array}\right] \label{eq:state_space_model}
\end{aligned}
\end{equation}
where $\mathbf{\Phi}_i \in \{0, 1\}$ represents the $i^{th}$ end-effector contact state at a given time instant and $\bm{r}_{1:n} =\{\bm{r}_{1}, \cdots, \bm{r}_{n} \}$. $[\mathbf{a}]_{\times} \in so(3)$ denotes the skew symmertic matrix spanned from a vector $\mathbf{a}\in \mathbb{R}^{3}$. Note that the matrix $\mathbf{B}$ explicitly depends on $\mathbf{I}$. In a similar fashion to~\cite{DiCarlo2018DynamicControl}, we perform an Euler step to obtain the discrete-time, time-varying linear system. The discrete-time dynamics are expressed as follows: 
\begin{align}
    \bm{x}_{k+1} = \mathbf{A}_k(\bm{x}_k)\bm{x}_k + \mathbf{B}_k(\bm{x}_k, \mathbf{I}_k, \bm{r}_{{1:n},k} )\mathbf{\Phi}_{{1:n},k}\bm{u}_k + \bm{d}_k \label{eq:discrete_time_dynamics}  
\end{align} 
where $\mathbf{A}_{k}(\bm{x}_{k}) = \Delta t \mathcal{A}_{k}(\bm{x}_{k})$ and $\mathbf{B}_{k}(\bm{x}_{k}) = \Delta t \mathcal{B}_{k}(\bm{x}_{k})$ with
\begin{equation*}
    \begin{aligned}
        \mathcal{A}_k(\bm{x}_k) &= \left[ 
                        \begin{array}{cccc}
                           \bm{1}_{3\times3}  & \bm{0}_{3\times3} & \mathbf{T}^{-1}(\bm{\Theta}_{\textrm{wbo},k})& \bm{0}_{3\times3} \\
                            \bm{0}_{3\times3} & \bm{1}_{3\times3} & \bm{0}_{3\times3} & \bm{1}_{3\times3}\\
                            \bm{0}_{3\times3} & \bm{0}_{3\times3} & \bm{1}_{3\times3} & \bm{0}_{3\times3}\\
                            \bm{0}_{3\times3}& \bm{0}_{3\times3} & \bm{0}_{3\times3} & \bm{1}_{3\times3}
                        \end{array}
                        \right], \\
        \mathcal{B}_k(\bm{x}_k) &= \left[ 
                        \begin{array}{cccc}
                           \bm{0}_{3\times3}  & \bm{0}_{3\times3} & \bm{0}_{3\times3} & \bm{0}_{3\times3} \\
                            \bm{0}_{3\times3} & \bm{0}_{3\times3} & \bm{0}_{3\times3} & \bm{0}_{3\times3}\\
                            \mathbf{I}^{-1}_k &  \mathbf{I}^{-1}_k [\bm{r}_{1, k}]_{\times} &  \mathbf{I}^{-1}_k &  \mathbf{I}^{-1}_k [\bm{r}_{2, k}]_{\times}\\
                            \bm{0}_{3\times3}& \bm{1}_{3\times3} / m & \bm{0}_{3\times3} & \bm{1}_{3\times3} / m
                        \end{array}
                        \right],\\                
        \mathbf{d}_k &= \left[
                        \begin{array}{cccc}
                            \bm{0}_{1\times3} & \bm{0}_{1\times3} & \bm{0}_{1\times3} & \mathbf{g}^{\top}\Delta t_k  
                        \end{array}
                        \right]^{\top}.
    \end{aligned}
\end{equation*}

In other words, given the reference trajectories $\bm{x}_k^{\textrm{ref}}, \mathbf{I}_k^{\textrm{ref}}$, $\bm{r}_{i,k}^{\textrm{ref}}$, and $\bm{r}_{i,k}^{\textrm{ref}}$ we approximate the linear and discrete system matrices $\mathbf{A}_k$ and $\mathbf{B}_k$ over the MPC prediction horizon.
\begin{assumption}
        The important assumption in our formulation is that the robot should follow the reference trajectory generated in the higher-level module (e.g., gait command or footstep planner). This assumption is reasonable since the MPC controller runs at a relatively high frequency ($\sim$$200$$\si{Hz}$). 
\end{assumption}
\subsection{Centroidal Composite Inertia Network}
In order to calculate and predict the (joint configuration-dependent) rotational inertia in~\eqref{eq:discrete_time_dynamics} throughout the MPC horizon, we need full-body joint reference trajectories. In practice, this requires additional modules, such as an offline whole-body trajectory optimizer~\cite{Zhou2022Momentum-AwareLocomotion}. However, we circumvent the need for such modules by leveraging the centroidal composite inertia network, similar to our previous work~\cite{ahn2021towr+}. 

We pre-train a CCINN that approximates the centroidal inertia tensor relative to the robot's base frame as a function of end-effectors' SE(3) relative to the base. This network is described as follows:
\begin{align} \label{eq:inertia_network}
        [I_{xx} \, I_{xy} \, I_{xz} \, I_{yy} \, I_{yz} \, I_{zz}]^T
 = \mathbf{I}_b(\bm{\varphi}_{1} \cdots, \bm{\varphi}_{n_\textrm{ee}})
\end{align}
where $\bm{\varphi}_{i} = (\bm{r}_{i,\{B\}}, \bm{\theta}_{i,\{B\}})$ denotes a pair of the position $\bm{r}_{i,\{B\}}$ and Euler angles $\bm{\theta}_{i,\{B\}}$ of the $i$-th end-effector with respect to the base frame of the robot. $\mathbf{I}_b: \mathbb{R}^{6\times n_{\textrm{ee}}} \mapsto \mathbb{R}^6$ is a mapping from the states of the end-effectors to the composite rigid-body inertia properties. In~\cite{ahn2021towr+}, we proposed a way to obtain the composite rigid-body inertia in a computationally efficient way. 


For efficient training, we collect data offline from a repetition of two main motions: forward steps and in-place turning, both parameterized accordingly. First, we carefully choose the upper and lower bounds of the feet and base in SE(3) for each motion, as well as the range of swing heights. Then, we uniformly sample initial and final points in SE(3) for each frame at each step and interpolate them to create motion trajectories. For each interpolated data point (input to the network), we solve the inverse kinematics (IK) to compute the joint configurations and centroidal inertia tensors (output to the network). 

The benefits of this NN are threefold in our framework. First, it integrates seamlessly with the SRB model, where only the task space coordinates (e.g., CoM and foot positions) are included in its states. Additionally, the NN facilitates fast online computation without the need to solve an expensive IK problem, as it allows full-body inverse kinematics to be solved offline. Lastly, its parametrization with the end-effectors allows our SRB-based MPC to accurately capture variable centroidal inertia by considering swing foot trajectories, an aspect often overlooked in ROMs.  


\subsection{Reference Trajectory Generation} \label{sec:reference_trajectory}
\subsubsection{Contact Sequence and Footholds}
In this work, we employ the periodic phase-based gait scheduler~\cite{Bledt2018ContactTerrains} for contact sequence generation and reference contact state for each leg, $\mathbf{\Phi}_{i, k}$ in~\eqref{eq:discrete_time_dynamics}. For a leg in the swing phase, we compute its landing foothold using Raibert heuristics~\cite{Raibert1983StableLocomotion} with a capture point-based~\cite{Pratt2012Capturability-basedHumanoid} feedback term and a centrifugal compensation term for turning: 
\begin{align} \label{eq:foot_pos_cmd}
    \bm{p}_{f,i}^{\textrm{cmd}} = \bm{p}_c + \kappa_{1} \dot{\bm{p}}_c + \kappa_{2}( \dot{\bm{p}}_c - \dot{\bm{p}}_c^{\textrm{cmd}}) + \kappa_{3} \dot{\bm{p}}_c \times \bm{\omega}^{\textrm{cmd}} 
\end{align}
where $\kappa_{1}= \frac{\Delta t_\textrm{rem}}{2}$, $\kappa_{2} = k_{1}$, and $\kappa_{3} = k_{2}\frac{h}{g}$ with the feedback gain $k_1$ and $k_{2}$.
In addition, $\bm{p}_{f,i} \in \mathbb{R}^2$ denotes the $i^{\textrm{th}}$ foot $xy$-position, $\Delta t_\textrm{rem} \in \mathbb{R}$ is the time remaining to the next stance change, and $h$ is a CoM height. With the computed landing foothold, a swing leg trajectory is generated using a cubic Bezier curve.
Additionally, the landing foot yaw orientation is designed with the torso heading and desired yaw velocity command: 
\begin{align}\label{eq:foot_ori_cmd}
    \mathbf{R}_{f,i}^{(\textrm{y})} = \textrm{Rot}_{z} \left(\frac{\omega^{\textrm{cmd}}_{\textrm{yaw}}\Delta t_\textrm{rem}}{2}\right)\mathbf{R}_{\textrm{torso}}^{(\textrm{y})}
\end{align}
where $\mathbf{R}^{(\textrm{y})}$ represents the yaw component of the rotation matrix. $\mathbf{R}_{f,i}$ and $\mathbf{R}_{\textrm{torso}}$ are the rotation matrices for the $i$-th foot and torso, respectively. In addition, $\textrm{Rot}_{z}(\theta)$ is the rotation matrix by an angle $\theta$ about the $z$ axis. Using the computed desired foot SE(3) from equations \eqref{eq:foot_pos_cmd} and \eqref{eq:foot_ori_cmd}, along with the predefined gait schedule, we generate the reference swing foot trajectory over the MPC horizon.


\subsubsection{CoM \& WBO} \label{sec:com_wbo_ref_traj} 
To provide the reference trajectory used in the cost function and dynamics constraints within the MPC problem, we compute the WBO ($\bm{\Theta}$) and CoM ($\bm{p}_c$) position references using Euler integration as follows: 
\begin{align}
   \bm{\Theta}_\textrm{yaw}^{\textrm{ref}} &= \bm{\Theta}_\textrm{yaw} + \Delta t \dot{\bm{\Theta}}_\textrm{yaw}^{\textrm{cmd}}, \\
   \bm{p}_c^{\textrm{ref}} &= \bm{p}_c + \Delta t\textrm{Rot}_z(\Theta_\textrm{yaw}^{\textrm{ref}})\dot{\bm{p}}_{c, \textrm{body}}^{\textrm{cmd}}, 
\end{align}
where $\dot{\Theta}_\textrm{yaw}^{\textrm{cmd}}$ and $\dot{\bm{p}}_{c, \textrm{body}}^{\textrm{cmd}}$ represent the yaw rate and the CoM velocity in the body frame commands, respectively. 

\subsubsection{Centroidal Rotational Inertia}
To account for the effect of varying centroidal rotational inertia over the MPC prediction horizon, we compute the reference trajectory of centroidal rotational inertia by utilizing the reference trajectories of the base and feet SE(3) as outlined in~\eqref{eq:inertia_network}. We then apply this trajectory to enforce the dynamics constraints specified in~\eqref{eq:discrete_time_dynamics}. Note that we generate the base reference trajectory similar to the method described in~\ref{sec:com_wbo_ref_traj} but with the initial base states at each MPC iteration.  

\subsection{Convex Model Predictive Control}
Our MPC aims to find the desired ground reaction wrenches which enable the SRB model with variable inertia to follow a given reference trajectory. In our formulation, the contact sequence is predetermined by the gait scheduler and footstep planner, as discussed in the subsequent subsection. This facilitates a convex MPC formulation, thereby enabling real-time performance. The MPC is formulated over a finite horizon $N$ as follows:
\begin{align}
   \min_{\substack{\bm{X}_{[s:s+N]|s}\\ \bm{U}_{[s:s+N-1]|s}}}& \quad 
   \ell_{f} (\bm{x}_{s+N|s}) + \sum_{k=0}^{N-1} \ell ( \bm{x}_{s+k|s}, \bm{u}_{s+k|s})\label{eq:mpc_cost}\\
   \textrm{subject to} 
   & \quad \bm{x}_{s+k+1|s} = \hat{\mathbf{A}}_{s+k|s}\bm{x}_{s+k|s} \nonumber\\
   & \qquad \qquad \quad  + \hat{\mathbf{B}}_{s+k|s}\bm{u}_{s+k|s}+ \bm{d}_{s+k|s}, \label{eq:mpc_dynamics}\\ 
   & \quad \underline{\bm{c}}_{s+k|s} \leq \mathbf{C}_{s+k|s}\bm{u}_{s+k|s} \leq \overline{\bm{c}}_{s+k|s} \label{eq:mpc_inequality_constrait},\\
   & \quad \forall k \in \{0, 1, \cdots, N-1\} \nonumber
\end{align}
with 
\begin{align*}
       \ell(\bm{x}_{s+k|s}, \bm{u}_{s+k|s}) &= \lVert \bm{x}_{s+k|s} - \bm{x}_{s+k|s}^\textrm{ref} \rVert^2_{\mathbf{M}_k} + \lVert \bm{u}_{s+k|s} \rVert^2_{\mathbf{N}_k},\\
       \ell_{f} (\bm{x}_{s+N|s}) &= \lVert \bm{x}_{s+N|s} - \bm{x}_{s+N|s}^\textrm{ref} \rVert^2_{\mathbf{M}_N}
\end{align*}
where $\bm{X}_{[s:s+N]|s} = [\bm{x}_{s|s}^{\top}, \bm{x}_{s+1|s}^{\top}, \cdots, \bm{x}_{s+N|s}^{\top}]^{\top}$ and $\bm{U}_{[s:s+N-1]|s} = [\bm{u}_{s|s}^{\top}, \bm{u}_{s+1|s}^{\top}, \cdots, \bm{u}_{s+N-1|s}^{\top}]^{\top}$ represent the optimal state and input sequence, respectively. $\bm{x}_{s+k|s}$ and $\bm{u}_{s+k|s}$ represent the state and input at time step $s+k$ predicted at time step $s$, respectively. In particular, $\bm{x}_{s|s}$ means the measured state at time step $s$. $\mathbf{M}_k$ and $\mathbf{N}_k$ are diagonal positive semidefinite matrices of weights.

~\eqref{eq:mpc_cost} describes the objective function of MPC to steer the state trajectory $\bm{X}_{[s:s+N]|s}$ close to the reference trajectory $\bm{X}^{\textrm{ref}}_{[s:s+N]|s}$, while minimizing control input sequence $\bm{U}_{[s:s+N-1]|s}$.~\eqref{eq:mpc_dynamics} presents the equality constraint on the dynamics model, where $\hat{\mathbf{A}}_{s+k|s}$ and $\hat{\mathbf{B}}_{s+k|s}$ are the approximate matrices of $\mathbf{A}_k$ and $\mathbf{B}_k$ in~\eqref{eq:discrete_time_dynamics}, respectively, using reference trajectory computed in the previous subsection~\ref{sec:reference_trajectory}.~\eqref{eq:mpc_inequality_constrait} specifies inequality constraints on the contact stability, incorporating unilateral constraint, contact wrench cone (CWC) constraint~\cite{caron2015contact}, and maximum reaction force constraint in the normal direction. 

This MPC formulation is rewritten in the form of a quadratic program (QP) and solved using the open-source solver HPIPM~\cite{Frison2020HPIPM:Control}, which leverages the sparsity and structure of the problem. After solving the QP, only the optimal solution $\bm{u}_0$ is utilized and passed to the low-level whole-body controller. 

\subsection{Low-level Control}
As the MPC optimizes the CoM, WBO, and contact wrench at the stance foot and swing foot positions given by the footstep planner, a whole-body controller (WBC) is needed to track those multiple tasks at a relatively higher frequency. The primary objective of WBC is to find the optimal joint torque commands while minimizing the given task error by leveraging the FBD in~\eqref{eq:multi-body dynamics} and the contact constraints. We extend the WBC in~\cite{Kim2019HighlyControl} to be compatible with the proposed MPC and DRACO 3 robot.

To compute the joint position, velocity, and acceleration while strictly maintaining task priority, we utilize inverse kinematics using the null-space projection technique but also consider the inclusion of internal constraints. We follow the method described in~\cite{Kim2019HighlyControl}, but we employ the null space of internal constraint Jacobian $\bm{N}_{\textrm{int}}(\mathbf{q})$ to perform the recursive iterations:
\begin{align}
    \mathbf{N}_0(\mathbf{q}) &= \mathbf{I} - \mathbf{J}_{c|\textrm{int}}(\mathbf{q})^{\dag}\mathbf{J}_{c|\textrm{int}}(\mathbf{q}), \\
    \ddot{\mathbf{q}}_0^{\textrm{cmd}} &= \overline{\mathbf{J}_{c|\textrm{int}}^{\textrm{dyn}}(\mathbf{q})}\left(-\dot{\mathbf{J}_c}(\mathbf{q}, \dot{\mathbf{q}})\dot{\mathbf{q}}\right),
\end{align}
where
\begin{equation}
\begin{aligned}
    \mathbf{J}_{c|\textrm{int}}(\mathbf{q}) &= \mathbf{J}_c(\mathbf{q}) \mathbf{N}_{\textrm{int}}(\mathbf{q}), \\
    \mathbf{N}_{\textrm{int}}(\mathbf{q}) &= \mathbf{I} - \mathbf{J}_{\textrm{int}}(\mathbf{q})^{\dag}\mathbf{J}_{\textrm{int}}(\mathbf{q}). 
\end{aligned}
\end{equation}
where $(\cdot)^{\dag}$ denotes an SVD-based pseudo-inverse. For the simplicity, we represent $\mathbf{J}(\mathbf{q})$ and $\mathbf{N}(\mathbf{q})$ as $\mathbf{J}$ and $\mathbf{N}$, respectively. $\overline{\mathbf{J}} = \mathbf{M}^{-1}\mathbf{J}^{\top}\left(\mathbf{J}\mathbf{M}^{-1}\mathbf{J}^{\top}\right)^{\dag}$ represents a dynamically consistent pseudo-inverse. This implies we compute the commands by prioritizing the internal constraint over the contact kinematic constraints.

Next, we formulate an optimization problem in the form of a QP to simultaneously optimize the desired joint acceleration and the reaction wrench while satisfying several constraints, including the FBD of the robot, contact constraints, and actuator limits. The formulation of the QP is written as below:
\begin{align}
    \min_{\ddot{\mathbf{q}}, \mathbf{F}_r, \bm{\ddot{x}}_c} \quad & \left\|\ddot{\mathbf{q}}^{\textrm{cmd}} - \ddot{\mathbf{q}}\right\|_{\mathbf{W}_{\ddot{\mathbf{q}}}}^2 + \left\|\mathbf{F}_r^{\textrm{MPC}} - \mathbf{F}_r\right\|_{\mathbf{W}_{\mathbf{F}_r}}^2 + \left\|\bm{\ddot{x}}_c\right\|_{\mathbf{W}_{\ddot{\bm{x}}_c}}^2  \nonumber \\
    \text{s.t.} \quad & \mathbf{S}_{f} \left(\mathbf{M} \ddot{\mathbf{q}} + \mathbf{b} + \mathbf{g} - \mathbf{J}_c^{\top} \mathbf{F}_r\right) = \mathbf{0}, \nonumber \\
    & \mathbf{J}_{\textrm{int}} \ddot{\mathbf{q}} + \dot{\mathbf{J}}_{\textrm{int}} \dot{\mathbf{q}} = \mathbf{0}, \nonumber\\
    & \bm{\ddot{x}}_c = \mathbf{J}_{c|\textrm{int}}\ddot{\mathbf{q}} + \dot{\mathbf{J}}_c\mathbf{\dot{q}},\\
    & \mathbf{U} \mathbf{F}_r \geq \mathbf{0}, \quad \mathbf{S}_{r} \mathbf{F}_r \leq \mathbf{F}_r^{\textrm{max}}, \nonumber\\
    & \bm{\tau}_{\text{min}} \leq \overline{\mathbf{S}_{a}\mathbf{N}_{\textrm{int}}}^{\top}\left(\mathbf{M} \ddot{\mathbf{q}} + \mathbf{N}_{\textrm{int}}^{\top}\left(\mathbf{b} + \mathbf{g}\right) \right. &  \nonumber\\
    & \left. \qquad \qquad -\left(\mathbf{J}_c\mathbf{N}_{\textrm{int}}\right)^{\top} \mathbf{F}_r \right) \leq \bm{\tau}_{\textrm{max}}, \nonumber
\end{align}
where $\mathbf{W}_{\ddot{\mathbf{q}}}$, $\mathbf{W}_{\mathbf{F}_r}$, and $\mathbf{W}_{\ddot{\bm{x}}_c}$ represent the diagonal weighting matrices for the joint acceleration errors, reaction wrench errors, and contact accelerations. $\mathbf{F}_r^{\textrm{MPC}}$ is the contact wrenches obtained from the proposed MPC. Compared to~\cite{Kim2019HighlyControl}, we newly add the contact acceleration penalization term since the joint acceleration relaxation can lead to the violation of contact constraints. After obtaining the optimal joint accelerations ($\mathbf{\ddot{q}}^{\star}$) and contact wrench ($\mathbf{F}_r^{\star}$) from the QP, the torque command is computed via inverse dynamics.

\begin{figure}[t]
    \centering
    \includegraphics[width=\columnwidth]{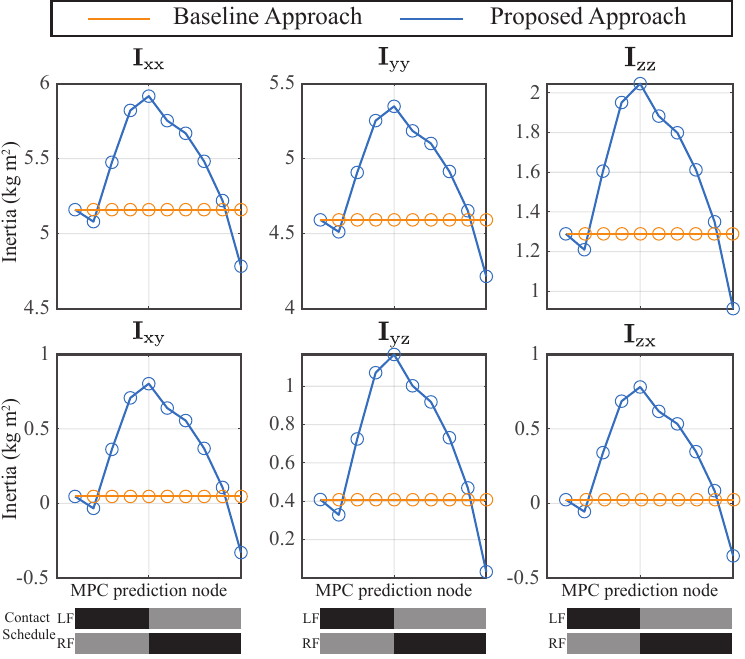}
    \caption{\textbf{Centroidal Inertia Prediction.} Using the CCINN, 6-dimensional centroidal rotational inertia tensors (in body frame) are predicted online across the MPC horizon (10 nodes) during forward walking at a speed of 1.2 m/s. The contact schedule is depicted with black for the contact phase and silver for the swing phase.}
    \label{fig:inertia_trajectory}
    \vspace{-5mm}
\end{figure}


\section{EXPERIMENTAL RESULTS} 
\label{sec:experimental_results}
\label{sec:experiment_results}
\subsection{Experimental Setup}

%


We performed simulations on the DRACO 3 humanoid robot~\cite{Bang2023ControlBody} in PyBullet~\cite{Coumans2016PyBulletLearning}. The robot weighs $39~\si{kg}$ and features rolling contact joints in its lower body, enabling a broad spectrum of movements. With a total of $n_a=25$ DoF, DRACO $3$ includes $6$ DoF per leg, $6$ DoF per arm, and $1$ DoF in the neck, as shown in~\cite{Bang2023ControlBody}. All Experiments were performed on a desktop PC with $48$GB of memory and an Intel\textsuperscript{\textregistered} Core\textsuperscript{TM} i7-7700K 4-core CPU @ 4.20GHz. All control modules employed in this paper are written in C++. The WBC control commands are executed at a frequency of 800~\si{Hz}, while the MPC controller operates at 200~\si{Hz}. For the execution of contact wrench commands from the MPC, we utilize a Zero-Order Hold (ZOH) between control signals. The MPC prediction horizon consists of 10 knot points, corresponding to 0.5~\si{s}.

The centroidal composite inertia network is composed of a 3-layer neural network consisting of a 12-dimensional input layer and two hidden layers (each with 64 neurons and a $\tanh$ activation function). To collect more reliable data for the NN training, we ensure that no self-collision occurs at any interpolated data point. This is done using Pinocchio's HPP-FCL library on our simplified collision model of DRACO $3$ involving cubes, spheres, and capsule primitives, for which a self-collision matrix was also configured. Moreover, we also verify that a valid IK solution exists at all instances given the rolling contact joint constraint of the legs of DRACO $3$. 

DRACO 3's WBO function approximator is constructed using monomial basis functions up to the 2nd order, with coefficients optimized using the algorithm described in~\cite{ChenIntegrableRobots} through IPOPT~\cite{Wachter2006OnProgramming}. We simplified the configuration space to 16 dimensions by excluding the ankle, wrist, and neck joints from the optimization, due to their minimal contribution to the CAM. The neural network and WBO function approximator were initially developed using the CasADi symbolic framework~\cite{Andersson2019CasADi:Control} in Python and subsequently exported to C++ using CasADi's internal code generation tool.

\begin{figure}[t]
    \centering
    \includegraphics[width=0.9\columnwidth]{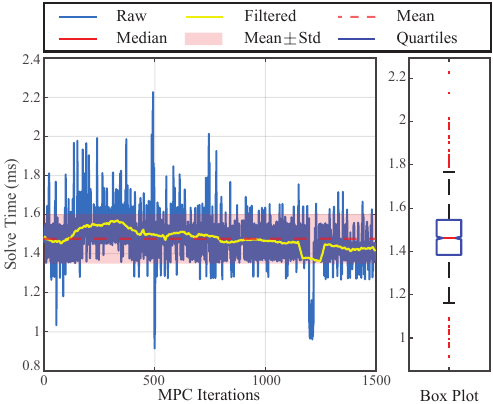}
    \vspace{-2mm}
    \caption{\textbf{MPC solve time.} The proposed MPC solve time is illustrated. The solve time is mostly below 2~\si{ms}.}
    \label{fig:mpc_solve_time}
    \vspace{-2mm}
\end{figure}

To evaluate the performance of our algorithm, we present three experimental cases across various dynamic locomotion scenarios: walking forward, walking diagonally, turning in place, and walking in a circular path. In each scenario, the robot was commanded to traverse flat terrain at varying commanded velocities, during which we measured the maximum velocity it could maintain without losing balance. The state-of-the-art SRB-based MPC locomotion controller~\cite{Kim2019HighlyControl, Li2021Force-and-moment-basedRobots} serves as a benchmark. For all test cases, the same low-level WBC with identical task setup and gains was employed. 

\subsection{Simulation Results} \label{sec:simulation_results}
To investigate the effect of centroidal rotational inertia in MPC, we compared our proposed approach against two baseline approaches with different inertia update rules: (1) using CCINN to predict the varying inertia over the MPC horizon (proposed approach), (2) employing constant inertia at a nominal configuration throughout the MPC horizon (baseline 1), and (3) using constant inertia over the MPC horizon but updating it at each MPC iteration (baseline 2), as depicted in Fig.~\ref{fig:inertia_trajectory}. Thanks to the convex formulation and the efficacy of CCINN, the average MPC solve time was approximately $1.4969$~\si{ms}, as shown in Fig.~\ref{fig:mpc_solve_time}. For the baseline approaches, we initialized the MPC with the torso's orientation and velocity at every MPC update. For our proposed approach, we initialized the MPC with the WBO and WBO velocity.

\begin{figure}[t!]
    \centering
    \includegraphics[width=\columnwidth]{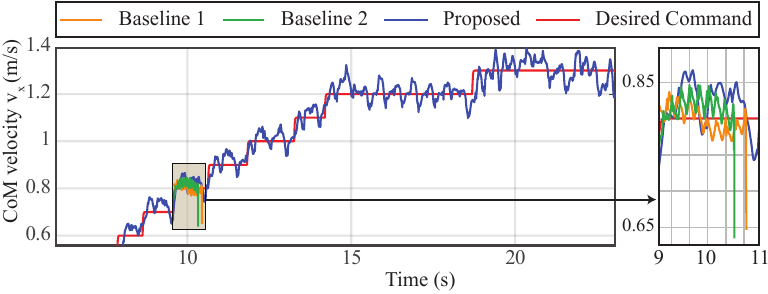}
    \vspace{-6mm}
    \caption{\textbf{Forward Walking Velocity Tracking Performance.} Case 1 employs nominal inertia at a nominal configuration. Case 2 uses constant inertia over the MPC horizon and updates it at every iteration. Case 3 showcases the proposed method.}
    \label{fig:velocity_tracking}
    \vspace{-5mm}
\end{figure}

\begin{figure}[t!]
    \centering
    \includegraphics[width=\columnwidth]{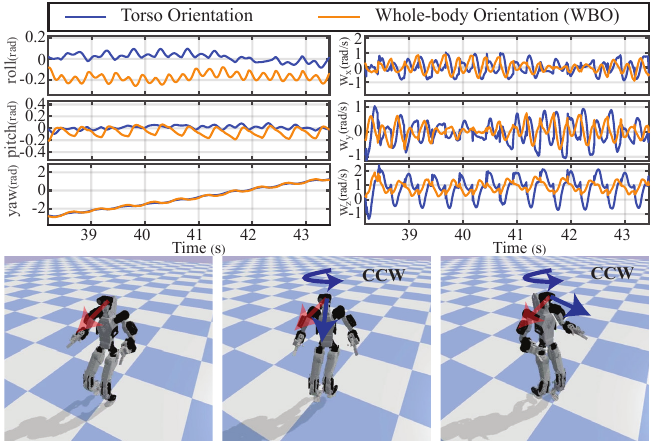}
    \vspace{-5mm}
    \caption{\textbf{Torso Orientation vs. WBO.} Comparison of torso orientation and WBO in the circular walking scenario (0.9~\si{m/s} and 0.9~\si{rad/s}). Red arrows represent the initial heading, and the blue arrows indicate the current heading with velocity commands.}
    \label{fig:torso_vs_wbo}
    \vspace{-5mm}
\end{figure}

In the forward-walking scenario, the proposed method successfully tracked reference velocities up to $1.3$\si{m/s}, outperforming the other two baseline methods, which only managed to track up to $0.7$\si{m/s}, as shown in Fig~\ref{fig:velocity_tracking}. In the diagonal walking scenario, the proposed method achieved tracking velocities of $0.9$~\si{m/s} in both the x and y directions, whereas the baseline methods were limited to $0.6$~\si{m/s} in both directions. In the turning in-place scenario, our method attained a turning rate of 2~\si{rad/s}, surpassing the baseline methods, which tracked up to 1.7~\si{rad/s}. In the circular walking scenario, our method achieved $0.9$~\si{m/s} and $0.9$~\si{rad/s}, while the baselines were constrained to $0.6$~\si{m/s} and $0.6$~\si{rad/s}. Notably, the inertia network precisely predicted the varying inertia over the MPC prediction horizon without the need for solving computationally expensive IK online. Furthermore, initializing the MPC with the WBO and WBO velocity enabled the SRB model to accurately capture the angular states of the FBD model. In particular, the difference between the torso orientation and WBO is illustrated in Fig.~\ref{fig:torso_vs_wbo}. With these advancements, the proposed MPC reduced the model discrepancy between the SRB and FBD models, enabling the generation of more dynamically consistent contact wrench references for WBC.

\section{CONCLUSIONS}
\label{sec:conclusion}
In this work, we introduced a MPC framework for agile and robust bipedal locomotion. By leveraging the variable centroidal inertia model and WBO, we developed a VI-MPC framework that generates dynamically consistent contact wrenches. We have demonstrated the efficacy of our approach across various locomotion scenarios, where the robot achieved increased step lengths, walking speeds, and turning speeds. Future work will focus on extensive hardware experiments with DRACO $3$ to further validate the proposed method. Additionally, we will explore the application of this framework to diverse humanoid loco-manipulation tasks that involve substantial changes in inertia.

\addtolength{\textheight}{-12cm}   




\section*{ACKNOWLEDGMENT}
This work was supported by the Office of Naval Research (ONR), Award No. N00014-22-1-2204.

\addtolength{\textheight}{12cm}
\bibliographystyle{IEEEtran}
\balance
\bibliography{references}

\end{document}